\documentclass[pdflatex,sn-mathphys-num]{sn-jnl}

\usepackage{graphicx}%
\usepackage{multirow}%
\usepackage{amsmath,amssymb,amsfonts}%
\usepackage{amsthm}%
\usepackage{mathrsfs}%
\usepackage[title]{appendix}%
\usepackage{xcolor}%
\usepackage{textcomp}%
\usepackage{manyfoot}%
\usepackage{booktabs}%
\usepackage{algorithm}%
\usepackage{algorithmicx}%
\usepackage{algpseudocode}%
\usepackage{listings}%

\theoremstyle{thmstyleone}%
\theoremstyle{thmstyletwo}%

\theoremstyle{thmstylethree}%

\begin{document}

\title[Article Title]{Text Over Image: Auditing Multimodal Robustness in Synthetic Medical Image Detection}







\author[1]{\fnm{Ching-Hao} \sur{Chiu}}\email{cchiu3@nd.edu}

\author[1]{\fnm{Hao-Wei} \sur{Chung}}\email{hchung6@nd.edu}

\author[1]{\fnm{Gelei} \sur{Xu}}\email{gxu4@nd.edu}

\author[1]{\fnm{Xueyang} \sur{Li}}\email{xli34@nd.edu}

\author[2]{\fnm{Pin-Yu} \sur{Chen}}\email{Pin-Yu.Chen@ibm.com}

\author[3,4]{\fnm{John} \sur{Kheir}}\email{John.Kheir@cardio.chboston.org}

\author[5]{\fnm{Meysam} \sur{Ghaffari}}\email{seyed\_ghaffaridehkordi@optum.com}

\author[5]{\fnm{Carlos} \sur{Morato}}\email{carlos.morato@optum.com}

\author[1]{\fnm{Ahmed} \sur{Abbasi}}\email{aabbasi@nd.edu}

\author*[1]{\fnm{Yiyu} \sur{Shi}}\email{yshi4@nd.edu}

\affil[1]{\orgname{University of Notre Dame}, \orgaddress{\city{Notre Dame}, \state{IN}, \country{USA}}}

\affil[2]{\orgname{IBM Research}, \orgaddress{\city{Yorktown Heights}, \state{NY}, \country{USA}}}

\affil[3]{\orgdiv{Department of Cardiology}, \orgname{Boston Children's Hospital}, \orgaddress{\city{Boston}, \state{MA}, \country{USA}}}

\affil[4]{\orgdiv{Department of Pediatrics}, \orgname{Harvard Medical School}, \orgaddress{\city{Boston}, \state{MA}, \country{USA}}}

\affil[5]{\orgname{Optum AI, UnitedHealth Group}, \orgaddress{\city{Minneapolis}, \state{MN}, \country{USA}}}


\abstract{With the rapid adoption of generative AI, synthetic medical images pose growing risks, including diagnostic deception and insurance fraud. Although prior work has explored vision–language model (VLM)–based synthetic image detection, these evaluations typically consider images in isolation. In clinical practice, however, images are interpreted alongside structured records and metadata, and VLMs are increasingly deployed under joint image–record inputs. We uncover a previously underexamined multimodal vulnerability: when given both modalities, VLMs may overweight record context in authenticity judgments, such that the same image receives different predictions solely due to changes in its accompanying text. This raises concerns about robustness in real-world deployment. To systematically characterize this effect, we reformulate synthetic medical image detection as an audit of multimodal robustness at the image–record interface and introduce a paired benchmark that holds the image fixed while swapping controlled metadata variants. Across multiple imaging modalities, we evaluate diverse open-weight and frontier API VLMs and find that changing the metadata context alone can flip authenticity judgments, with accuracy on authentic images dropping by 61.1\% on average under an explicit AI-origin tag. We further propose an inference-time mitigation pipeline that detects and neutralizes provenance shortcuts without model retraining, substantially outperforming direct prompt-based suppression on the affected subset. Our benchmark provides a standardized tool for assessing and improving multimodal robustness beyond image-only settings. Code and data will be released upon acceptance.}




\renewcommand{\thefootnote}{}%
\footnotetext{This paper is an extended version of our work accepted at MICCAI 2026~\cite{chiu2026beyond}. Compared with the conference version, the present manuscript adds additional provenance perturbations, paired statistical analysis, mitigation experiments and broader discussion of deployment implications for multimodal medical AI.}%
\renewcommand{\thefootnote}{\arabic{footnote}}%

\maketitle

\section{Introduction}\label{sec1}

Every day, medical images support high-stakes clinical and administrative decisions, and these workflows are increasingly mediated by AI systems~\cite{ryu2025vision, xu2026comprehensive}. In this setting, vision-language models (VLMs) are increasingly integrated into workflows that combine visual inputs with textual records~\cite{nath2025vila,sellergren2025medgemma,xu2025lingshu,chen2024huatuogpt}, such as report generation~\cite{hartsock2024vision,pan2025medvlm} and insurance claims pipelines~\cite{cheng2026hybrid}. However, rapid advances in generative imaging~\cite{reddy2024generative,konz2024anatomically} have reshaped the risk landscape. The increasing quality and scale of synthetic medical images pose practical threats to healthcare systems~\cite{khosravi2025exploring,kondylakis2025review}, as highly realistic forgeries can mislead even medical experts~\cite{li2025toward} and enable diagnostic deception or medical insurance fraud. In insurance claims pipelines, for example, a forged image that is mistakenly trusted may enable substantial improper payouts, while a legitimate image that is wrongly flagged may trigger prolonged investigations and delayed treatment. 
This raises a deployment-level challenge: as VLMs increasingly mediate high-stakes workflows, how robust are their decisions to AI-generated medical images that may enter the pipeline as inputs?

Although synthetic medical image detection has been widely studied, it is still predominantly evaluated with image-only inputs~\cite{li2025toward,grabovski2025back,ali2025enhancing}, treating visual evidence as the primary signal of authenticity. As a result, existing evaluations cannot fully characterize multimodal decision behavior in healthcare deployment, where medical images are often judged alongside clinical records and authenticity decisions are formed under joint image–record inputs.

\begin{figure}[!t]
  \centering
  \includegraphics[width=\linewidth]{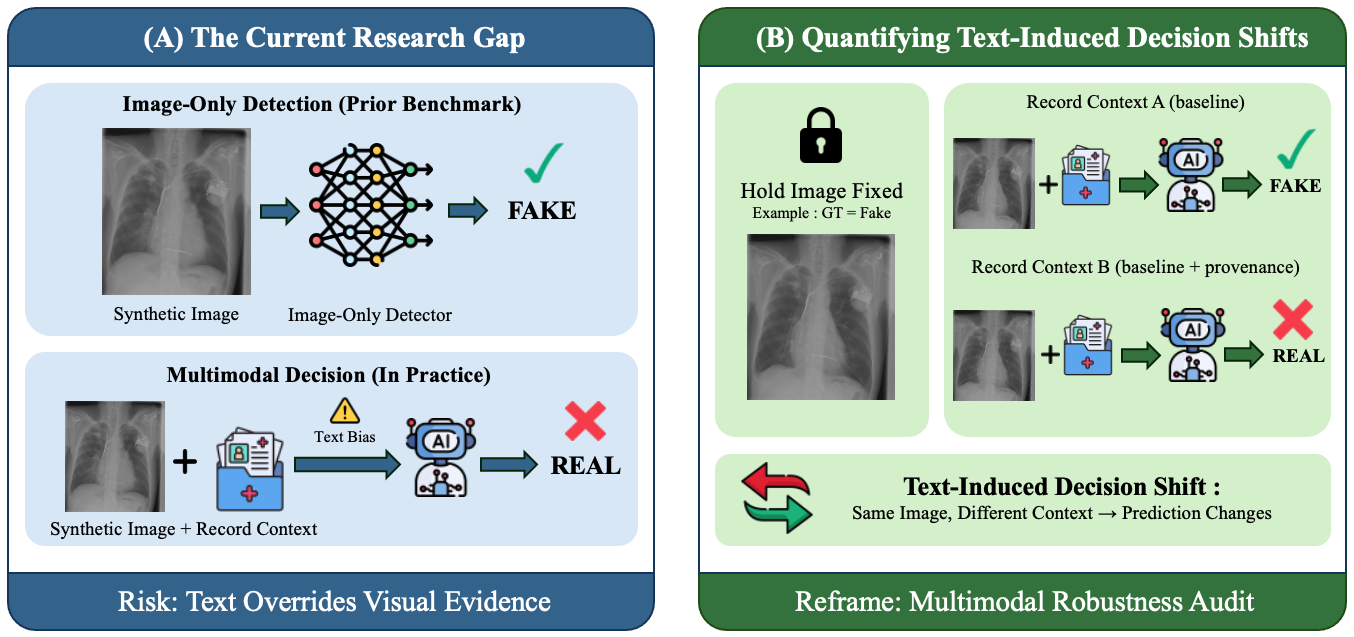} 
\caption{\textbf{Research concept: auditing multimodal robustness for synthetic medical image detection.}
\textbf{(A)} Prior benchmarks often evaluate authenticity from images alone, while real-world decisions use multimodal inputs where accompanying records can influence the verdict.
\textbf{(B)} We measure \textbf{Text-Induced Decision Shift} by holding the image fixed and changing only the accompanying record (represented as structured metadata) with a controlled provenance field.}
  \label{fig:concept}
\end{figure}

This gap matters because a key deployment risk is not merely detecting whether an image is synthetic, but whether accompanying records can bias or steer the authenticity judgment for the same image. We refer to changes in authenticity judgment induced by altering only the accompanying record context as a \textbf{Text-Induced Decision Shift}. Recent work on VLMs shows that holding the image fixed while perturbing the accompanying text can induce substantial decision changes, and models may overly trust textual cues even when they contradict clear visual evidence~\cite{deng2025words,zhang2026images}. One might argue that an image-only detector could serve as an initial filter. However, even with such filtering, errors are unavoidable and many real workflows still require joint image-record decisions; therefore, multimodal behavior at the image-record interface must be evaluated directly. This raises our central research question: for the same image, how sensitive is a VLM's authenticity judgment to the accompanying record context, and to what extent can record context steer the final verdict when the image is unchanged? Fig.~\ref{fig:concept} summarizes this gap and previews our paired framework for quantifying text-induced decision shift.

To quantify deployment risk under joint image-record inputs, we introduce a 
benchmark to audit multimodal robustness in synthetic medical image detection 
across NIH Chest X-ray14~\cite{wang2017chestx} (NIH-CXR14), ISIC2019~\cite{tschandl2018ham10000}, 
and a private in-house pediatric chest X-ray dataset (PediCXR), spanning open-weight 
VLMs and frontier API models (e.g., GPT-5~\cite{singh2025openai}). This benchmark is grounded in a real deployment context, where 
AI-assisted documentation is entering clinical workflows and provenance information 
is increasingly visible to downstream systems~\cite{tai2024ai,stults2025evaluation}, 
creating conditions where a single provenance field can steer authenticity judgments 
in high-stakes decisions. As a result, we instantiate record context 
as structured metadata variants that differ only in a single provenance field. 
Since real-world provenance information is diverse and difficult to control, 
rather than replicating real-world metadata diversity, we design a controlled 
\texttt{Source} field as a reproducible probe, and keep the image fixed to 
isolate whether provenance cues alone can influence authenticity judgments 
despite unchanged visual evidence. To systematically expose this failure mode 
across different directions and signal strengths, we define three provenance 
conditions: \texttt{Source: Hospital} (toward-real), \texttt{Source: AI-preprocessed} 
(ambiguous toward-fake), and \texttt{Source: Edited by Nano Banana (AI Editing 
Technique)} (explicit toward-fake, representing an upper-bound stress test; 
referred to as \texttt{AI-edited}). We find that changing only this field can flip authenticity judgments; for example, on authentic images, adding \texttt{Source: AI-edited} reduces accuracy 
by \textbf{61.1\%} on average across all models and datasets. To address this 
vulnerability, we further propose \textbf{Shortcut-Aware Visual Commitment (SAVC)}, 
a four-stage inference-time pipeline that automatically detects and neutralizes 
provenance shortcuts without model retraining. For example, in the subset where text-induced decision shift has occurred, SAVC recovers up to \textbf{99.8\%} accuracy on MedGemma-27B \cite{sellergren2025medgemma}, an open-weight medical VLM, substantially 
outperforming prompt-based suppression (e.g., instructing the model to ignore 
the \texttt{Source} field), which remains largely ineffective with accuracy as low as \textbf{1.8\%}. This work makes three contributions:

\begin{itemize}
\item \textbf{Benchmark method.} We introduce a paired, controllable benchmark 
that audits multimodal robustness in synthetic medical image detection by holding 
the image fixed while varying a single provenance field across three conditions 
spanning different directions and signal strengths. This design quantifies 
text-induced decision shift and isolates how metadata context can override visual 
evidence in VLM-based authenticity judgments.

\item \textbf{Cross-model finding.} Across multiple datasets and model families, text-induced decision shift is pervasive for both authentic and synthetic images, including in frontier API models (e.g., GPT-5).

\item \textbf{Inference-time mitigation.} We propose SAVC, a four-stage inference-time pipeline that detects, confirms, and neutralizes provenance shortcuts without model retraining, evaluated on affected cases where text-induced decision shifts occur. Unlike prompt-based suppression, SAVC requires no prior knowledge of which fields are problematic, and uses counterfactual verification to empirically confirm 
shortcuts before neutralizing them.

\end{itemize}

\section{Method}\label{sec2}

\subsection{Synthetic Data Generation Pipeline}
\label{subsec:synthetic-data-generation}
We generate paired synthetic images using an LLM-guided edit-verify-refine loop~\cite{chen2025med} with quality control, converting each authentic image $x$ with label $y$ to a sampled target label $y' \neq y$ while preserving medical plausibility and consistency with the original acquisition context. For each dataset, we construct a target pool $\mathcal{Y}$ using label frequencies calculated from the official evaluation split. For ISIC2019 (multi-class), $\mathcal{Y}$ contains all diagnostic classes. For multi-label datasets (NIH-CXR14 and PediCXR), $\mathcal{Y}$ includes individual labels and frequent co-occurring label sets. Given an original label $y$, we sample $y'\neq y$ from $\mathcal{Y}$ in proportion to its label frequency.

For each $(x, y, y')$ triplet, we run an editing pipeline. We use Gemini-2.5-Pro~\cite{comanici2025gemini} to produce a structured, reusable instruction template and to revise it in a consistent format, and use Gemini-2.5-Flash-Image~\cite{comanici2025gemini} for image editing. Each edited image is then verified by Gemini-2.5-Pro via an LLM-as-judge procedure. We accept an edit only if the judge confirms (i) the target diagnosis is present, (ii) anatomy is plausible, and (iii) the image appears realistic; otherwise, we revise the instruction using structured failure feedback and retry for up to five rounds. Finally, all accepted edits undergo final review by two in-house clinicians (one for CXR, one for dermoscopy); we retain an image only if it passes verification for plausibility, realism, and consistency with the target diagnosis.

We form \textbf{Base Metadata} by retaining essential demographics and diagnostic finding labels from the original dataset metadata, and update the label field to $y'$ for synthetic samples. Starting from this Base Metadata, we create three \textbf{Source-Augmented Metadata} variants (\textbf{Source-H}, \textbf{Source-AI-prep} and \textbf{Source-AI-Nano Metadata}) by appending a controlled \texttt{Source} field. This single-field \texttt{Source} intervention yields a controlled, task-aligned test for multimodal authenticity judgments, isolating the impact of provenance cues.

\subsection{Evaluation Framework}
\label{subsec:evaluation-protocol}

\subsubsection{Task Formulation and Input Conditions.}
We formulate the task as synthetic medical image detection, where VLMs must determine whether a medical image is \textbf{Real} (authentic) or \textbf{Fake} (synthetic). We evaluate each model under five input conditions: \textbf{I-Only} (image only), \textbf{I+Base} (image + Base Metadata), \textbf{I+Source-H} (image + Source-H Metadata), \textbf{I+Source-AI-prep} (image + Source-AI-prep Metadata), and \textbf{I+Source-AI-Nano} (image + Source-AI-Nano Metadata). To quantify text-induced decision shift, we perform paired judgments that hold the image fixed while swapping only the metadata context between matched variants of the same image.

\subsubsection{Prompt and Output Format.} 

We use a standardized forensic prompting template to ensure comparability across different architectures. Each model completes a checklist of five static visual criteria, including texture, noise patterns, edges, anatomical plausibility, and color, followed by a check for potential contradictions (within the image for image-only inputs; between visual evidence and the accompanying metadata when multimodal inputs are provided). The metadata 
is placed after the task description under multimodal conditions. We do not convert checklist responses into a decision; instead, the model outputs the final decision by weighing all available evidence. After the checklist, we enforce a dual verdict output format: the model outputs a \texttt{FINAL ANSWER} (a holistic judgment using all provided inputs) followed by a \texttt{VISUAL VERDICT} (a judgment based on the image alone). We use \texttt{FINAL ANSWER} as the primary output for evaluation and treat 
\texttt{VISUAL VERDICT} as a same-pass consistency probe that distinguishes two failure modes: (1) \texttt{FINAL ANSWER} shifts while \texttt{VISUAL VERDICT} does not, indicating the provenance cue overrides the integrated verdict without affecting image-based reasoning; and (2) both verdicts shift together, suggesting metadata also rewrites visual reasoning. 

\subsection{Mitigation: Shortcut-Aware Visual Commitment}
\label{subsec:mitigation}

To address text-induced decision shift at inference time, we propose 
\textbf{Shortcut-Aware Visual Commitment (SAVC)}, a four-stage inference-time 
pipeline that identifies and neutralizes harmful provenance shortcuts without 
model retraining. Algorithm~\ref{alg:savc} summarizes the full procedure.

\begin{algorithm}[t]
\caption{Shortcut-Aware Visual Commitment (SAVC)}
\label{alg:savc}
\begin{algorithmic}[1]
\Require Image $x$, metadata $m$, target VLM $\mathcal{M}$, orchestrator LLM $\mathcal{O}$
\Ensure Final authenticity decision $v^*$

\State \textit{// Stage 0a: Shortcut Detection}
\State $\mathcal{S} \leftarrow \mathcal{O}(m)$ \Comment{flag fields that alone could determine Real/Fake}
\If{$\mathcal{S} = \emptyset$} skip to Stage~1 \EndIf

\State \textit{// Stage 0b: Counterfactual Verification}
\State $v_\text{orig} \leftarrow \mathcal{M}(x, m)$
\State $v_\text{cf} \leftarrow \mathcal{M}(x, m')$, where $m' = m$ with $\mathcal{S}$ replaced by \texttt{"Unknown"}
\State confirmed $\leftarrow (v_\text{cf} \neq v_\text{orig})$

\State \textit{// Stage 1: Visual Forensics}
\State $(v_\text{vis}, c_\text{vis}) \leftarrow \mathcal{M}(x)$ \Comment{image-only; $c_\text{vis} \in \{$High, Med, Low$\}$}

\State \textit{// Stage 2: Deterministic Integration}
\If{not confirmed} $v^* \leftarrow v_\text{vis}$
\ElsIf{$c_\text{vis} = $ High} $v^* \leftarrow v_\text{vis}$
\Else\ $v^* \leftarrow v_\text{cf}$
\EndIf
\State \Return $v^*$
\end{algorithmic}
\end{algorithm}

\paragraph{Stage 0a: Shortcut Detection.}
An orchestrator LLM inspects the metadata alone and flags fields whose values 
could steer an authenticity judgment without any visual evidence, annotating 
each with an inferred direction (Toward-Fake or Toward-Real). 
No predefined field list is required, allowing SAVC to adapt to task-dependent metadata schemas in which different fields may serve as shortcuts.

\paragraph{Stage 0b: Counterfactual Verification.}
Each flagged field is replaced with ``Unknown'' and the target VLM is queried 
twice using the same standardized forensic prompt as the benchmark evaluation---once 
with the original metadata ($v_\text{orig}$) and once with the neutralized version 
($v_\text{cf}$). Reusing the benchmark prompt ensures that any observed verdict 
change is attributable solely to the provenance signal, rather than to differences 
in prompt framing. We replace the field value rather than removing the field 
entirely, so that the only difference between the two inputs is the provenance 
signal itself, enabling clean causal attribution. A shortcut is confirmed only 
when $v_\text{cf} \neq v_\text{orig}$, providing empirical causal evidence 
rather than heuristic detection alone.

\paragraph{Stage 1: Visual Forensics.}
The target VLM analyzes the image without any metadata using a prompt that 
shares the same visual checklist as the benchmark evaluation, with an additional 
request for a confidence level $c_\text{vis} \in \{$High, Medium, Low$\}$ 
alongside the verdict $v_\text{vis}$. Running this stage independently of any metadata ensures that $v_\text{vis}$ 
serves as an uncontaminated visual baseline, free from any textual influence.

\paragraph{Stage 2: Deterministic Integration.}
If no shortcut is confirmed, the pipeline defaults to $v_\text{vis}$ to avoid 
introducing unverified metadata influence into the final decision. When a shortcut 
is confirmed and $c_\text{vis}$ is high, $v_\text{vis}$ is used directly, as 
strong visual evidence is sufficient without metadata supplementation. When a 
shortcut is confirmed but visual confidence is medium or low, $v_\text{cf}$ is 
used instead, as the counterfactual-cleaned verdict still leverages useful 
clinical context while being free of the identified shortcut. This design ensures that identified shortcuts are neutralized before metadata 
is allowed to influence the final decision.

\section{Results}\label{sec3}

\subsection{Experimental Setup and Datasets}
\label{subsec:exp_setup_models}

To ensure that our robustness evaluation generalizes across medical domains, we construct a multimodal benchmark encompassing distinct visual semantics. We use NIH-CXR14~\cite{wang2017chestx}, where synthetic findings can blend into grayscale anatomy and appear highly realistic, and ISIC2019~\cite{tschandl2018ham10000}, where edits in images with rich texture often leave visible artifacts. We choose these established datasets because they provide official splits and structured metadata, enabling reproducible experiments and controlled counterfactual metadata contexts. We also include a private in-house dataset, PediCXR, to reduce concerns that certain models may have had prior exposure to public benchmark test data. The PediCXR dataset was used under institutional IRB approval.

In terms of sample selection, we sequentially iterate through the source test sets and retain cases that successfully pass the synthetic generation pipeline. For the public benchmarks (NIH-CXR14 and ISIC2019), we retain the first 500 valid samples, and for PediCXR, we retain the first 200. The pipeline discards on average 1.2 (NIH-CXR14), 0.1 (ISIC2019), and 0.4 (PediCXR) failed attempts per retained image. Following Sec.~\ref{subsec:synthetic-data-generation}, we construct paired \textbf{Real}--\textbf{Fake} images, where \textbf{Real} is the original image and \textbf{Fake} is its AI-edited counterpart. For each paired image, we evaluate the image alone (I-Only) and pair it with four matched metadata variants (Base, Source-AI-prep, Source-AI-Nano, Source-H) as described in Sec.~\ref{subsec:evaluation-protocol}, using each resulting input to obtain the model’s authenticity judgment.

To analyze robustness across model types, we group evaluated systems into four families: open-weight medical VLMs, open-weight general VLMs, frontier API models, and DetectFake VLMs, i.e., VLMs explicitly designed for synthetic image detection (we include general domain DetectFake VLMs as baselines, since medical synthetic image detectors are often image-only and do not accept record text, making them incompatible with our multimodal setting). Medical VLMs include MedGemma~\cite{sellergren2025medgemma}, Lingshu~\cite{xu2025lingshu}, and HuatuoGPT-Vision~\cite{chen2024huatuogpt}, which are trained on biomedical corpora covering CXR and dermoscopy; general VLMs include InternVL-3.5~\cite{wang2025internvl3}, Qwen3-VL~\cite{bai2025qwen3}, and Llama-3.2-Vision~\cite{grattafiori2024llama}. To investigate capacity effects, we compare different model capacities within the same architecture: MedGemma (4B vs.\ 27B), Lingshu (7B vs.\ 32B), Qwen3-VL (8B vs.\ 32B), and InternVL-3.5 (8B vs.\ 14B). All open-weight models are run on a compute node equipped with four NVIDIA A10 GPUs; due to resource constraints, we evaluate HuatuoGPT-Vision-7B and Llama-3.2-Vision-11B. For frontier API models, we evaluate GPT-5~\cite{singh2025openai}, GPT-5-mini, Gemini-2.5-Flash~\cite{comanici2025gemini}, Gemini-2.5-Pro, and Gemini-3-Flash. For DetectFake VLMs, we evaluate FakeVLM~\cite{wen2025spot} and FakeShield~\cite{xu2024fakeshield}. All models use default decoding settings from each official implementation or API.

\begin{figure*}[p]
  \centering
  \includegraphics[width=\linewidth,height=0.99\textheight,keepaspectratio]{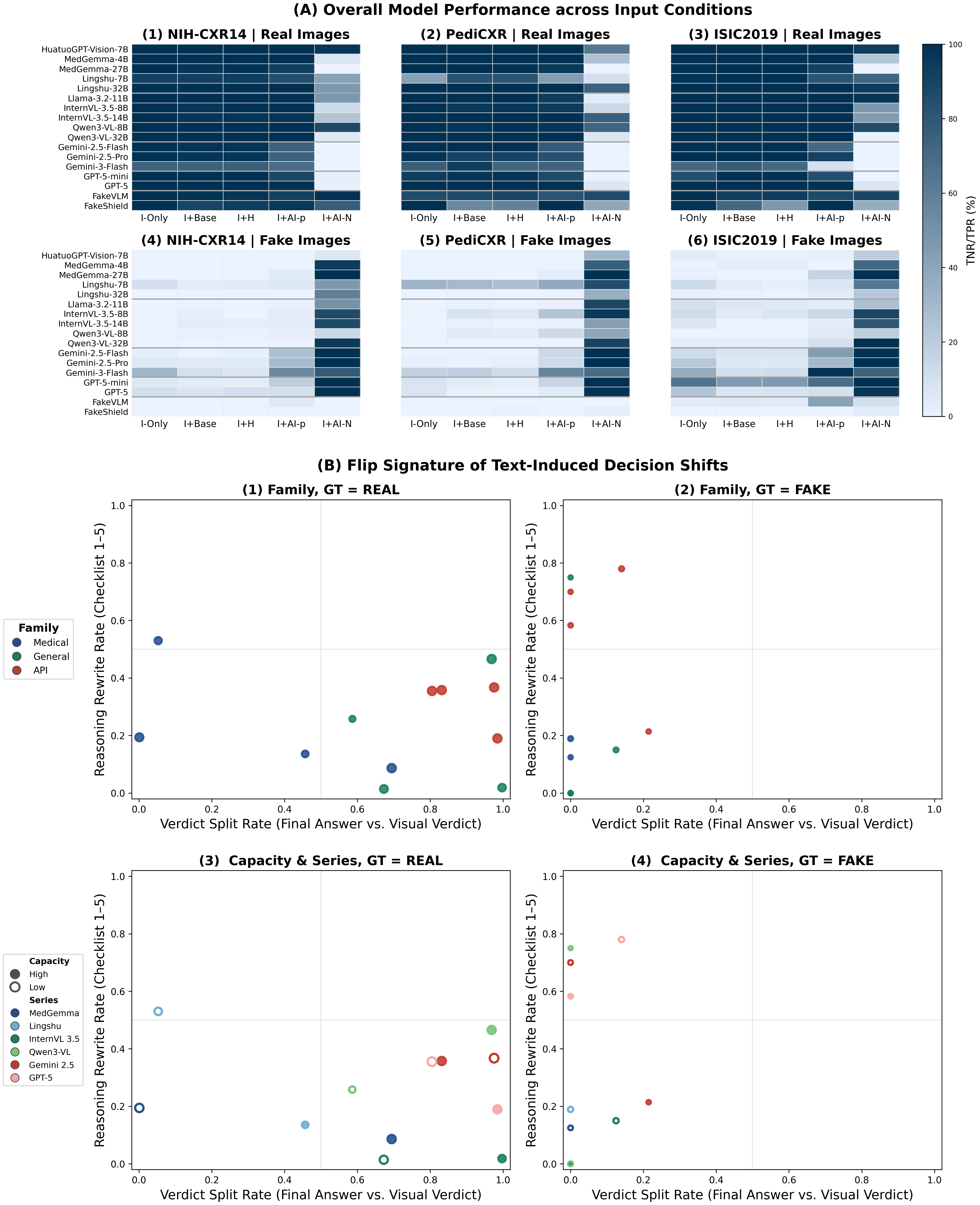} 
\caption{\textbf{Main results: text-induced decision shift and flip signatures.}
\textbf{(A)} TNR (True Negative Rate) and TPR (True Positive Rate) on NIH-CXR14, PediCXR, and ISIC2019 under five input conditions: I-Only, I+Base, I+Source-H (I+H), I+Source-AI-prep (I+AI-p), and I+Source-AI-Nano (I+AI-N).
\textbf{(B)} Flip signature scatter over flip cases (aggregated across datasets), with point size proportional to the number of flip cases. The rates are computed over flip cases and reflect failure modes rather than overall performance.}
  \label{fig:main_results_overall_accuracy}
\end{figure*}

\subsection{Main Results}
\label{sec:main_results}
\subsubsection{Text-Induced Decision Shift and Within-Family Differences.}

Fig.~\ref{fig:main_results_overall_accuracy}(A) summarizes text-induced decision shift under five input conditions. We report TNR (True Negative Rate) for Real images and TPR (True Positive Rate) for Fake images across NIH-CXR14, PediCXR, and ISIC2019 (invalid outputs counted as incorrect), with models grouped by family.

On Real images (Fig.~\ref{fig:main_results_overall_accuracy}(A)(1--3)), most models perform well under I-Only and remain strong under I+Base. However, appending an AI-origin provenance field to the metadata can sharply reduce accuracy by pushing models to label authentic images as Fake, with severity scaling with signal strength: I+Source-AI-Nano produces the largest drops, while I+Source-AI-prep induces a weaker but still observable shift in many models, confirming that the vulnerability persists beyond explicit provenance cues. For example, on NIH-CXR14 (Real), switching from I+Base to I+Source-AI-Nano reduces MedGemma-27B from 97.6\% to 0.0\% and MedGemma-4B from 100.0\% to 7.6\%. Similar drops also appear in general VLMs, with severity varying across model series and capacity; capacity does not consistently predict robustness. For instance, Qwen3-VL-32B collapses across all three Real-image settings, whereas Qwen3-VL-8B remains substantially more robust. On the other hand, for InternVL-3.5, the 14B variant is more robust than the 8B variant on NIH-CXR14 and PediCXR, while their performance is comparable on ISIC2019. The same failure mode appears in frontier API models and DetectFake VLMs (e.g., Gemini models degrade to nearly zero under I+Source-AI-Nano), whereas others (e.g., FakeVLM) are comparatively more resistant. 

On Fake images (Fig.~\ref{fig:main_results_overall_accuracy}(A)(4--6)), many models have low I-Only accuracy, indicating that detecting highly realistic AI-edited medical images from the image alone is challenging. Adding Base Metadata does not help: I+Base often fails to improve over I-Only and may degrade performance, suggesting that metadata without provenance field can distract the model when it is visually uncertain. For example, on ISIC2019 (Fake), InternVL-3.5 drops from I-Only to I+Base (8B: 32.4\%$\rightarrow$10.0\%; 14B: 9.6\%$\rightarrow$0.6\%). In contrast, Source-AI-Nano Metadata often inflate accuracy on Fake images; given the weak I-Only and I+Base baselines, these apparent gains are best explained by reliance on the provenance field rather than improved visual detection of synthetic artifacts. I+Source-AI-prep shows a similar but weaker inflation effect, further supporting that the model is following the provenance signal rather than detecting visual artifacts. Source-H Metadata often reduces or maintains performance relative to Base Metadata (e.g., PediCXR (Fake): Gemini-3-Flash, 19.0\%$\rightarrow$12.0\%), suggesting that provenance field indicating a hospital source may increase false negatives when the model is uncertain about the visual evidence. DetectFake VLMs perform poorly across medical domains, likely due to domain mismatch from general domain training. Together, these patterns represent text-induced decision shift in both directions: provenance cues steer predictions regardless of visual evidence, with direction determined by the cue itself.

\subsubsection{Flip Signature Analysis of Text-Induced Decision Shift.}
Although Fig.~\ref{fig:main_results_overall_accuracy}(A) shows that the provenance field can flip correctness under a fixed image, it cannot explain the mechanism, i.e., do flips arise mainly from changes in the integrated decision while the image-based judgment remains stable, or do they also involve rewriting visual reasoning? We therefore analyze \textbf{flip case}s, where the same image’s \texttt{FINAL ANSWER} switches from correct to incorrect between (i) image + Base Metadata and (ii) image + Source-Augmented Metadata. To maximize the number of eligible flip cases, we focus on the conditions 
that produce the most harmful direction of shift for each ground-truth label: 
\textbf{I+Source-AI-Nano} for Real images and \textbf{I+Source-H} for Fake 
images, where the provenance cue pushes the verdict away from the correct answer. We summarize each model with two rates in Fig.~\ref{fig:main_results_overall_accuracy}(B). Using \texttt{VISUAL VERDICT} as an image-based same-pass consistency probe, the \textbf{Verdict Split Rate} is the fraction of flip cases where \texttt{FINAL ANSWER} $\neq$ \texttt{VISUAL VERDICT} under the Source-Augmented condition. We use a same-pass probe because two separate passes cannot reveal whether metadata shifts only the integrated decision or also rewrites the model's visual reasoning. Since \texttt{VISUAL VERDICT} is generated after the model has seen the metadata, it is not fully independent of textual influence (I-Only is the truly independent baseline); this makes our analysis conservative, as metadata that also shifts \texttt{VISUAL VERDICT} would lower the observed Verdict Split Rate and underestimate the true divergence. The \textbf{Reasoning Rewrite Rate} is the fraction of flip cases where the checklist changes between I+Base and the corresponding Source-Augmented input; after labeling each checklist item (1--5) as \{Positive, Negative, Neutral\} (supporting \texttt{REAL}, supporting \texttt{FAKE}, or uncertain), we count a rewrite if any item flips polarity across the two inputs. 

In Fig.~\ref{fig:main_results_overall_accuracy}(B)(1), for flip cases on Real images,  medical VLMs tend to show lower Verdict Split Rates, meaning \texttt{FINAL ANSWER} and \texttt{VISUAL VERDICT} more often move together when provenance changes. In contrast, general and frontier API models more often flip \texttt{FINAL ANSWER} while keeping \texttt{VISUAL VERDICT} unchanged, resulting in higher Verdict Split Rates. In the series \& capacity view (Fig.~\ref{fig:main_results_overall_accuracy}(B)(3)), compared to smaller models, larger models often exhibit higher Verdict Split but lower Reasoning Rewrite Rates. This suggests that although the integrated decision (\texttt{FINAL ANSWER}) is shifted by the provenance field, larger models can maintain their image-based reasoning and keep their \texttt{VISUAL VERDICT} relatively stable. For Fake images in Fig.~\ref{fig:main_results_overall_accuracy}(B)(2) and (B)(4), Verdict Split Rates are generally low compared to Real images, indicating that \texttt{FINAL ANSWER} and \texttt{VISUAL VERDICT} usually agree. The main cross-model variation occurs in Reasoning Rewrite, with API models often higher. Capacity effects are less consistent on Fake images. Notably, points for Fake image flip cases tend to be smaller, indicating fewer eligible flips. This is expected under our flip definition, because many models start with low baseline accuracy on Fake images, leaving fewer initially correct predictions that can be flipped by provenance field.



\begin{figure*}[t]
  \centering
  \includegraphics[width=\linewidth,height=0.99\textheight,keepaspectratio]{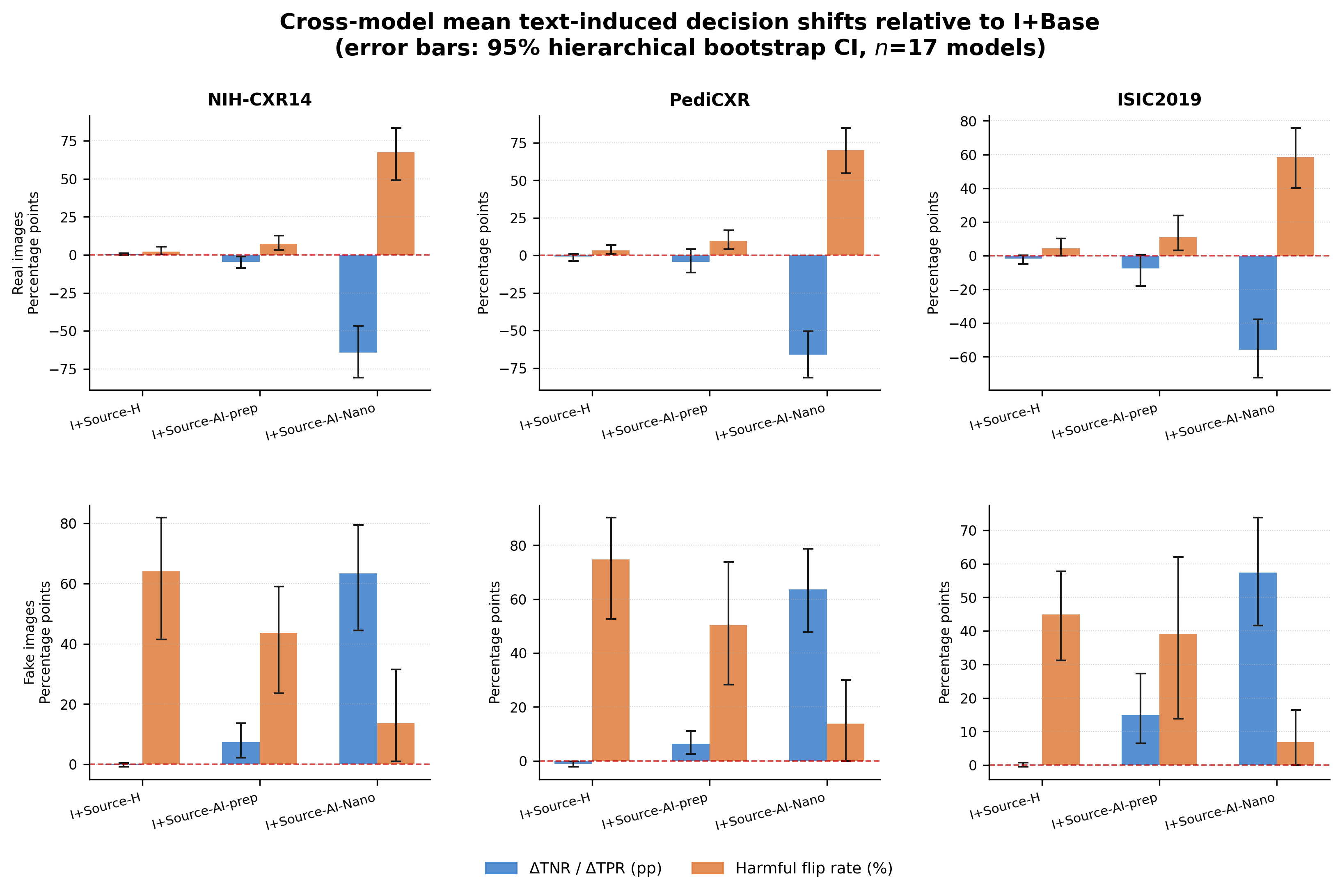} 
\caption{Cross-model mean text-induced decision shifts relative to I+Base, averaged across
    17 evaluated models. Each panel reports $\Delta$TNR (Real images, top row)
    or $\Delta$TPR (Fake images, bottom row) and the harmful flip rate for each source condition across three datasets. Error bars denote 95\% hierarchical bootstrap confidence intervals accounting for both model-level and image-level variability. Harmful flip rate is computed conditionally over cases correctly classified under I+Base.}
  \label{fig:effect_size}
\end{figure*}

\subsubsection{Quantifying Text-Induced Shifts with Paired Effect Sizes.}
\label{sec:effect_sizes}
While Fig.~\ref{fig:main_results_overall_accuracy}(A) provides a visual overview 
of how accuracy changes across models and conditions, heatmaps alone do not 
quantify how large these shifts are or how statistically robust they are across 
models and image samples. To address this, we compute paired effect sizes that 
directly measure how much each \texttt{Source} field changes model decisions, 
holding the image fixed and varying only the metadata context, and we report 
hierarchical bootstrap confidence intervals that account for both model-level and 
image-level variability.

For Real images, we report $\Delta\text{TNR}(c) = \text{TNR}(c) - \text{TNR}(\text{I+Base})$
for each source condition $c \in \{\text{I+Source-H},\ \text{I+Source-AI-prep},\ 
\text{I+Source-AI-Nano}\}$, capturing how much a given provenance field changes 
the rate of correctly identifying authentic images relative to the no-provenance 
baseline.
For Fake images, we report $\Delta\text{TPR}(c)$ analogously, measuring how much 
each source condition changes the rate of correctly identifying synthetic images.
We further report the \emph{harmful flip rate} for each condition $c$, defined as 
the fraction of cases correctly classified under I+Base but incorrectly classified 
under condition $c$, directly quantifying the decision harm induced by each 
provenance field.
For each dataset and source condition, point estimates are computed per model and 
then averaged across the 17 evaluated models. Uncertainty is estimated via 95\% 
hierarchical bootstrap confidence intervals (1000 iterations): in each iteration we 
resample the 17 models with replacement and, within each resampled model, resample 
its paired cases with replacement, recompute the per-model effect, and average across 
the sampled models; the 2.5th and 97.5th percentiles of the resulting distribution 
form the interval. This two-level scheme accounts for both model-level heterogeneity 
and image-level sampling variability~\cite{davison1997bootstrap}. The harmful flip 
rate is undefined when a model has no I+Base-correct cases (zero denominator) and is 
excluded from the corresponding average rather than set to zero. Results are 
summarized in Fig.~\ref{fig:effect_size}.

Fig.~\ref{fig:effect_size} reveals three consistent patterns across datasets. First, for Real images, I+Source-H shows little change relative to I+Base: both
$\Delta$TNR and the harmful flip rate are small, indicating that a hospital-origin provenance field alone does not
meaningfully shift authenticity judgments on authentic images. For Fake images, $\Delta$TPR under I+Source-H stays near zero with CIs crossing zero in NIH-CXR14 and ISIC2019, with a small negative shift in PediCXR; with many models already at low TPR baselines under
I+Base, further degradation from a hospital-origin cue is hard to detect at the
aggregate level. However, the harmful flip rate appears high (45.9--74.7\%), with
CIs entirely above zero. Since the harmful flip rate is computed over only the small subset
of cases correctly identified under I+Base, even a handful of verdict changes can produce a large percentage. Within this subset, a hospital-origin cue can still flip a model's verdict from Fake to Real, representing a concrete instance of text-induced decision shift. Second, I+Source-AI-prep produces weaker but directionally consistent shifts.
On Real images, $\Delta$TNR excludes zero on NIH-CXR14 and harmful flip rates have CIs above zero across all datasets, showing that even an ambiguous AI cue causes measurable harm.
On Fake images, I+Source-AI-prep shows positive $\Delta$TPR with CI excluding zero across all three datasets. This suggests that the ambiguous AI-preprocessing cue can shift some models toward Fake predictions, although the effect is weaker than for the explicit AI-origin cue.
Third, I+Source-AI-Nano produces the largest and most consistent effects.
Real-image TNR drops by 55.7--66.1\,pp on average, with CIs entirely below zero,
indicating that an explicit AI-origin cue introduces substantial false positive
risk for authentic images. This is mirrored by high harmful flip rates on Real
images with CIs entirely above zero, showing that many cases
correctly classified under I+Base are flipped to incorrect Fake predictions once
the explicit AI-origin cue is added.
Fake-image TPR rises by 59.4--63.6\,pp, consistent with the provenance-driven
pattern discussed above rather than improved visual detection.
Although CIs under I+Source-AI-Nano are wide, reflecting heterogeneity in model sensitivity, even the lower bounds remain far from zero, confirming that the
average effect is robust across the evaluated models and datasets.

Notably, a positive $\Delta$TPR and a positive harmful flip rate can co-occur on
Fake images: an AI-related cue raises overall Fake-image TPR,
yet a subset of cases correctly identified as Fake under I+Base are flipped to
Real once the cue is added. The two metrics are therefore complementary rather than contradictory: 
$\Delta$TPR captures the overall change in detection accuracy, while the harmful 
flip rate isolates the localized harm hidden beneath that overall gain. Even
when a provenance cue improves aggregate Fake-image detection, it does not do so
uniformly and can still overturn a fraction of previously correct verdicts.

\subsection{Mitigation Results}
\begin{table}[h]
\centering
\caption{SAVC mitigation accuracy on the affected subset (cases where I+Base is correct but I+Source-augmented condition is incorrect; baseline 
accuracy is 0\% by construction). Gain = SAVC Acc. $-$ Vis-only Acc.}
\label{tab:savc_accuracy}
\footnotesize
\setlength{\tabcolsep}{4pt}
\begin{tabular}{llcrcccc}
\toprule
Model & Dataset & Cond. & $n$ & Ig-S & Vis-only & SAVC & Gain \\
\midrule
MedGemma-27B & NIH-CXR14   & AI-Nano    & 488 & 17.2\% & 91.8\% & 99.2\%  & $+$7.4\% \\
MedGemma-27B & ISIC2019    & AI-Nano    & 500 & 1.8\%  & 92.6\% & 99.8\%  & $+$7.2\% \\
MedGemma-27B & PediCXR & AI-Nano    & 198 & 28.3\% & 96.5\% & 100.0\% & $+$3.5\% \\
\midrule
Lingshu-7B   & NIH-CXR14   & AI-prep & 44  & 11.4\% & 97.7\% & 97.7\%  & 0\% \\
Lingshu-7B   & ISIC2019    & AI-prep & 93  & 15.1\% & 100.0\% & 100.0\% & 0\% \\
Lingshu-7B   & PediCXR & AI-prep & 86  & 51.2\% & 81.4\% & 81.4\%  & 0\% \\
\bottomrule
\end{tabular}
\end{table}


We evaluate SAVC on a targeted subset that isolates cases where text-induced decision shift has occurred, i.e., cases where I+Base is correct but the Source-augmented condition is incorrect. This design directly assesses affected-case recovery rather than overall performance. Baseline accuracy on this subset is 0\% by construction. We evaluate two backbone models under two provenance conditions: MedGemma-27B under I+Source-AI-Nano, which exhibited the most severe text-induced decision shift in our benchmark, and 
Lingshu-7B under I+Source-AI-prep, a subtler provenance cue with ambiguous AI signal. GPT-5-mini serves as the orchestrator for shortcut detection 
in all experiments. As an ablation, we also evaluate prompt-based suppression, where the model is directly instructed to ignore the \texttt{Source} field, to test 
whether a simpler intervention can resolve the vulnerability.

Table~\ref{tab:savc_accuracy} reports the final accuracy of the full SAVC pipeline (i.e., the verdict $v^*$ after \textit{Stage~2} integration), the visual-only accuracy from SAVC's internal \textit{Stage~1} (Vis-only), and prompt-based suppression accuracy (Ig-S) on the affected subset. For MedGemma-27B under I+Source-AI-Nano, SAVC recovers 99.2--100.0\% accuracy across all three datasets, consistently 
outperforming both visual-only inference (91.8--96.5\%) and prompt-based suppression. 
Notably, Ig-S remains largely ineffective, with accuracy as low as 1.8\% on 
ISIC2019, confirming that directly instructing the model to ignore the \texttt{Source} field does not reliably neutralize its influence. For Lingshu-7B under I+Source-AI-prep, SAVC matches the accuracy of its own visual-only stage (Vis-only) while still outperforming Ig-S across all three datasets. Across both models, Ig-S remains largely ineffective despite the model being explicitly instructed to ignore the \texttt{Source} field. These results suggest that instruction-based suppression is not a reliable 
solution, as models may still be influenced by textual cues despite explicit instructions~\cite{deng2025words}. Rather than relying on the model to follow suppression instructions, SAVC first identifies and removes confirmed shortcuts from the input before the model makes its final judgment.

\begin{table}[h]
\centering
\caption{SAVC decision pathway distribution on the affected subset. 
Confirmed: fraction of cases where the shortcut is causally verified 
($v_\text{cf} \neq v_\text{orig}$) in \textit{Stage~0b}. 
Vis-pri.: shortcut detected in \textit{Stage~0a} but not confirmed in \textit{Stage~0b}; 
pipeline defaults to $v_\text{vis}$. 
Vis-high: shortcut confirmed and $c_\text{vis} = \textsc{High}$; 
pipeline uses $v_\text{vis}$ directly. 
CF-pri.: shortcut confirmed and $c_\text{vis} \neq \textsc{High}$; 
pipeline uses the shortcut-cleaned verdict $v_\text{cf}$. Note: Vis-high $+$ CF-pri. $=$ Confirmed; Confirmed $+$ Vis-pri. $= 100\%$.}
\label{tab:savc_decision}
\footnotesize
\setlength{\tabcolsep}{4pt}
\begin{tabular}{llccccc}
\toprule
Model & Dataset & Cond. & Confirmed & Vis-pri. & Vis-high & CF-pri. \\
\midrule
MedGemma-27B & NIH-CXR14   & AI-Nano    & 94.1\% & 5.9\%  & 81.6\% & 12.5\% \\
MedGemma-27B & ISIC2019    & AI-Nano    & 99.6\% & 0.4\%  & 77.0\% & 22.6\% \\
MedGemma-27B & PediCXR & AI-Nano    & 98.0\% & 2.0\%  & 89.4\% & 8.6\%  \\
\midrule
Lingshu-7B   & NIH-CXR14   & AI-prep & 18.2\% & 81.8\% & 18.2\% & 0\%    \\
Lingshu-7B   & ISIC2019    & AI-prep & 18.3\% & 81.7\% & 18.3\% & 0\%    \\
Lingshu-7B   & PediCXR & AI-prep & 19.8\% & 80.2\% & 19.8\% & 0\%    \\
\bottomrule
\end{tabular}
\end{table}
Table~\ref{tab:savc_decision} reveals the mechanism behind these two operating modes. Across all conditions and datasets, the orchestrator (GPT-5-mini) 
consistently identified the \texttt{Source} field as a candidate shortcut in \textit{Stage~0a}, confirming that semantic shortcut detection generalizes across both 
explicit and subtle provenance cues without a predefined field list. For MedGemma-27B, shortcut confirmation rates are high (94.1--99.6\%), 
indicating that the explicit AI-editing tag consistently produces a causal verdict 
flip in counterfactual testing. The CF-priority pathway is activated in 8.6--22.6\% 
of cases, accounting for the accuracy gain over visual-only inference as shown in Table~\ref{tab:savc_accuracy}. For 
Lingshu-7B, confirmation rates drop to 18.2--19.8\%, reflecting that the subtler 
\texttt{Source: AI-preprocessed} cue does not consistently flip the verdict in counterfactual 
testing. All confirmed cases are routed through the visual-high-confidence pathway, 
and the CF-priority pathway is absent, resulting in no gain beyond the visual 
baseline. This conservative behavior reflects a core design principle: when causal 
confirmation is unavailable, SAVC defaults to visual grounding rather than 
introducing unverified metadata influence.

\section{Discussion and Conclusion}
This work identifies a gap between how synthetic medical image detection is evaluated and how it is deployed: current evaluations treat it as an image-only visual forensics problem, whereas real deployments make decisions under joint image--record inputs. To close this gap, we recast the task in a multimodal image--record setting and audit the robustness of VLMs through a paired benchmark. Our paired benchmark shows that holding the image fixed and varying
only a single metadata field can flip authenticity judgments across diverse model
families, datasets, and provenance conditions. This suggests that image-only
evaluations may overestimate deployment reliability when VLMs operate under
joint image--record inputs.

On authentic images, an AI-origin provenance cue can sharply increase false
positives, causing models to label real images as synthetic despite unchanged
visual evidence. On synthetic images, the same cue can inflate the true positive
rate, but since many models perform poorly under image-only or I+Base conditions, this apparent improvement is more consistent with shortcut reliance
on the provenance field than with stronger visual forensics. In the opposite
direction, a hospital-origin cue can increase false negatives on synthetic images
by encouraging models to treat them as authentic. These patterns are not
confined to a single model family: text-induced decision shifts appear across
medical VLMs, general VLMs, frontier API models, and DetectFake VLMs, and
larger models are not consistently more robust. The three provenance conditions should be interpreted as a controlled spectrum that spans different strengths and directions of the signal, rather than a complete simulation of the diversity of real-world clinical documents. Future work should extend to more realistic settings, such as multi-field perturbations and clinical documents where provenance information appears implicitly rather than as an explicit labeled field.

SAVC is proposed as an inference-time recovery method for cases where
text-induced decision shift has already occurred. A natural baseline is
to instruct the model to ignore suspicious fields directly in the prompt.
However, this approach has two limitations: it requires prior knowledge
of which fields are problematic, making it impractical in deployment
settings where shortcuts are not known in advance, and VLMs may still
be influenced by textual cues despite explicit suppression instructions,
as evidenced by the low Ig-S accuracy in our experiments. SAVC addresses both issues: it requires no predefined field list, automatically identifying candidate shortcuts through semantic reasoning and confirming their causal effect through counterfactual
verification before neutralization. Nevertheless, two limitations of SAVC should be noted: (1)~SAVC adopts a conservative design: it neutralizes all confirmed shortcuts
even when they happen to provide correct signals, choosing to remove
potentially misleading cues rather than risk relying on them.
(2)~When visual confidence is not high, the pipeline uses the shortcut-cleaned verdict. If residual clinical context bias remains after shortcut neutralization, the final decision may still degrade, though this did not occur in our experiments. Beyond these limitations, future work should
explore more targeted algorithmic solutions, such as alignment tuning or
attention-based regularization, to complement inference-time approaches
like SAVC.

More broadly, the synthetic images reflect a specific generation pipeline and may contain generator-specific artifacts or selection bias toward edits that are easier to generate successfully. However, this limitation does not undermine our core paired analysis, since each comparison holds the image fixed and varies only the metadata context. In addition, prompt phrasing may influence multimodal integration despite our use of standardized templates, and we do not systematically evaluate prompt sensitivity. Investigating prompt robustness remains an important direction for future work. Overall, our results argue that robustness in synthetic medical image detection should be assessed at the image--record interface: a reliable model should not only detect visual artifacts, but also maintain stable authenticity judgments when record context changes.

\section*{Declarations}

M.G. and C.M. are Optum AI employees; Optum AI provided no funding. No other competing interests are declared.

\bibliography{sn-bibliography}

\end{document}